%% file: main.tex
\documentclass[sigconf, screen]{acmart}

\copyrightyear{2020}
\acmYear{2020}
\setcopyright{acmcopyright}\acmConference[MM '20]{Proceedings of the 28th ACM International Conference on Multimedia}{October 12--16, 2020}{Seattle, WA, USA}
\acmBooktitle{Proceedings of the 28th ACM International Conference on Multimedia (MM '20), October 12--16, 2020, Seattle, WA, USA}
\acmPrice{15.00}
\acmDOI{10.1145/3394171.3413570}
\acmISBN{978-1-4503-7988-5/20/10}

\acmSubmissionID{2486}

\input{macros}

\begin{document}

\title{Emotions Don't Lie: An Audio-Visual Deepfake Detection Method using Affective Cues}

\author{Trisha Mittal$^1$,  
Uttaran Bhattacharya$^1$, 
Rohan Chandra$^1$,  Aniket Bera$^1$, 
Dinesh Manocha$^{1,2}$}
% \email{dmanocha@umd.edu}
\orcid{1234-5678-9012}
\affiliation{%
  \institution{$^1$Department of Computer Science, University of Maryland, College Park, USA \\ $^{1,2}$Department of Electrical and Computer Engineering, University of Maryland, College Park, USA}
    % \institution{Dept. of ECE}
  % \streetaddress{P.O. Box÷ 1212}
  \city{\small{Project Page: \url{https://gamma.umd.edu/deepfakes/}}}
%   \state{Maryland}
}

% \author{Trisha Mittal$^1$,  
% Uttaran Bhattacharya$^1$, 
% Rohan Chandra$^1$,  Aniket Bera$^1$, 
% Dinesh Manocha$^{1,2}$}
% \institute{Department of Computer Science, University of Maryland, College Park, USA \\ Department of Electrical and Computer Engineering, University of Maryland, College Park, USA||
% \small{Project Page: \url{https://gamma.umd.edu/deepfakes/}}}


\input{sections/00-abstract}




% \includegraphics[width=\textwidth]{images/teaser-rohan.png}
%  \caption{\textit{Deepfake Detection: }We propose a deep learning-based method that identifies deepfake videos with $90\%$ average accuracy. To the best of our knowledge, ours is the first approach that combines audio and visual information from videos to identify the subject's emotional state (\textit{i.e.} affect), and classify a video as real or fake depending on extracted affect.} 
%   Fake multimedia content obtained from an online social platform~\cite{trump,obama}.}
% \vspace{-15pt}

\maketitle

\input{sections/01-introduction}

\input{sections/02-relatedwork}

\input{sections/03-approach}
\input{sections/04-implementation}
\input{sections/05-results}
\input{sections/06-conclusion}
\input{sections/07-acknowlwdgements}
% \begin{acks}
% \end{acks}

\bibliographystyle{ACM-Reference-Format}
\bibliography{acmart}


\end{document}

%% file: macros.tex
\usepackage{multicol,multirow}
\usepackage{enumitem}
\usepackage{subfig}
\usepackage{xcolor}
\usepackage{float}
\usepackage{colortbl}
\usepackage{graphicx}
\usepackage{comment}
\usepackage{caption}
\usepackage{amsmath,amssymb}
\usepackage{booktabs}
\usepackage{color}
\usepackage{float}
\newcommand{\fr}{f_\textrm{real}}
\newcommand{\sr}{s_\textrm{real}}
\newcommand{\ff}{f_\textrm{fake}}
\newcommand{\spf}{s_\textrm{fake}}
\newcommand\Tstrut{\rule{0pt}{2.6ex}}         % = `top' strut

%% file: sections/00-abstract.tex
\begin{abstract}
We present a learning-based method for detecting real and fake deepfake multimedia content.
%more specifically deepfake content. 
To maximize information for learning, we extract and analyze the similarity between the two audio and visual modalities from within the same video. Additionally, we extract and compare affective cues corresponding to perceived emotion from the two modalities within a video to infer whether the input video is \textit{``real''} or \textit{``fake''}. We propose a deep learning network, inspired by the Siamese network architecture and the triplet loss. To validate our model, we report the AUC metric on two large-scale deepfake detection datasets, DeepFake-TIMIT Dataset and DFDC. We compare our approach with several SOTA deepfake detection methods and report per-video AUC of $84.4\%$ on the DFDC and $96.6\%$ on the DF-TIMIT datasets, respectively. To the best of our knowledge, ours is the first approach that simultaneously exploits audio and video modalities and also perceived emotions from the two modalities for deepfake detection.
\end{abstract}

%% file: sections/01-introduction.tex
\section{Introduction}
\label{sec:introduction}
%%%%%%%%%%%%%%%%%%%%%%%%%%%%%%%%%%%%%%%%%%%%%%%%%%%%%%%%%%%%%%%%%%%%%%
Recent advances in computer vision and deep learning techniques have enabled the creation of sophisticated and compelling forged versions of social media images and videos~(also known as \textit{``deepfakes''})~\cite{FaceSwap,FakeApp,DeepFaceLab,DFaker,FaceSwap-GAN}.Due to the surge in AI-synthesized deepfake content, multiple attempts have been made to release benchmark datasets~\cite{DeepFake-TIMIT,faceforensics++,DFD,DFDC} and algorithms~\cite{two-stream,Mesonet,UADFV,FWA,VA-MLP,faceforensics++,MultiTask,Capsule,temporal1,temporal2,multimedia_forensics} for deepfake detection. DeepFake detection methods classify an input video or image as \textit{``real''} or  \textit{``fake''}. 
% one of either ``" mostly explored as a binary classification task, with inputs being images or videos and output labels being \textit{``real''} or  \textit{``fake''}. 
Prior methods exploit only a single modality, i.e., only the facial cues from these videos either by employing temporal features or by exploring the visual artifacts within frames. Other than these modalities, multimodal approaches have also exploited the contextual information in video data to detect fakes~\cite{video_verification_ontext_cues}. There are many other applications of video processing that use and combine multiple modalities for audio-visual speech recognition~\cite{ASR}, emotion recognition~\cite{m3er,cmu-mosei}, and language and vision tasks~\cite{visionandtext1,visionandtext2}. These applications show that combining multiple modalities can provide complementary information and lead to stronger inferences. Even for detecting deepfake content, we can extract many such modalities like facial cues, speech cues, background context, hand gestures, and body posture and orientation from a video. When combined, multiple cues or modalities can be used to detect whether a given video is real or fake. 

%%%%%%%%%%%%%%%%%%%%%%%%%%%%%%%%%%%%%%%%%%%%%%%%%%%%%%%%%%%%%%%%%%%%%%
In this paper, the key idea used for deepfake detection is to exploit the relationship between the visual and audio modalities extracted from the same video. Prior studies in both the psychology~\cite{psych1} literature, as well as the multimodal machine learning literature~\cite{multimodaltaxonomy} have shown evidence of a strong correlation between different modalities of the same subject~\cite{psych1}. More specifically,~\cite{neural,sync1,sync2,sync3} suggest some positive correlation between audio-visual modalities, which have been exploited for multimodal perceived emotion recognition. For instance, \cite{psych2,psych3} suggests that when different modalities are modeled and projected into a common space, they should point to similar affective cues. Affective cues are specific features that convey rich emotional and behavioral information to human observers and help them distinguish between different perceived emotions~\cite{affective_cues}. These affective cues comprise of various positional and movement features, such as dilation of the eye, raised eyebrows, volume, pace, and tone of the voice. We exploit this correlation between modalities and affective cues to classify \textit{``real''} and  \textit{``fake''} videos. 
\input{tables/dataset_details}
%%%%%%%%%%%%%%%%%%%%%%%%%%%%%%%%%%%%%%%%%%%%%%%%%%%%%%%%%%%%%%%%%%%%%%

\textbf{Main Contribution: } We present a novel approach that simultaneously exploits the audio (speech) and video (face) modalities and the perceived emotion features extracted from both the modalities to detect any falsification or alteration in the input video. To model this multimodal features and the perceived emotions, our learning method uses a Siamese network-based architecture. At training time, we pass a real video along with its deepfake through our network and obtain modality and perceived emotion embedding vectors for the face and speech of the subject. We use these embedding vectors to compute the triplet loss function to minimize the similarity between the modalities from the fake video and maximize the similarity between modalities for the real video. 

%%%%%%%%%%%%%%%%%%%%%%%%%%%%%%%%%%%%%%%%%%%%%%%%%%%%%%%%%%%%%%%%%%%%%%
The novel aspects of our work include: 
\begin{enumerate}
    \item We propose a deep learning approach to model the similarity (or dissimilarity) between the facial and speech modalities, extracted from the input video, to perform deepfake detection.
    % \item We propose a deep learning approach that combines the audio-visual information in a video to identify the emotional state of the subject in the video. to model the similarity (or dissimilarity) between the facial and speech modalities, extracted from the input video, to perform deepfake detection.
    
    \item We also exploit the affect information, \textit{i.e.}, perceived emotion cues from the two modalities to detect the similarity (or dissimilarity) between modality signals, and show that perceived emotion information helps in detecting deepfake content.
\end{enumerate}
%%%%%%%%%%%%%%%%%%%%%%%%%%%%%%%%%%%%%%%%%%%%%%%%%%%%%%%%%%%%%%%%%%%%%%
We validate our model on two benchmark deepfake detection datasets, DeepFakeTIMIT Dataset~\cite{DeepFake-TIMIT}, and DFDC~\cite{DFDC}. We report the Area Under Curve~(AUC) metric on the two datasets for our approach and compare with several prior works. We report the per-video AUC score of 84.4\%, which is an improvement of about 9\% over SOTA on DFDC, and our network performs at-par with prior methods on the DF-TIMIT dataset.

%% file: tables/dataset_details.tex
\begin{table*}[th]
    \centering
    \caption{\small{\textbf{Benchmark Datasets} for DeepFake Video Detection. Our approach is applicable to datasets that include the audio and visual modalities. Only two datasets~(highlighted in blue) satisfy that criteria and we evaluate the performance on those datasets. Further details in Section~\ref{subsec:used-datasets}.}}
    \label{tab:datasets}
    \resizebox{.9\textwidth}{!}{%
    \begin{tabular}{lcrrrcccc}
    \toprule
    \textbf{Dataset} & \textbf{Released} & \multicolumn{3}{c}{\textbf{\# Videos}} & \multicolumn{2}{c}{\textbf{Video Source}} & \multicolumn{2}{c}{\textbf{Modes}} \\
    \cmidrule{3-9}
    &  & \textbf{Real} & \textbf{Fake} & \textbf{Total} & \textbf{Real} & \textbf{Fake} & \textbf{Visual} & \textbf{Audio} \\
    \midrule
    UADFV~\cite{UADFV} & Nov 2018 & 49 & 49 & 98 & YouTube & FakeApp~\cite{FakeApp} & \checkmark & $\times$ \\
    % \midrule
    \rowcolor{blue!20} DF-TIMIT~\cite{DeepFake-TIMIT} & Dec 2018 & 0 & 620 & 620$^*$ & VidTIMIT~\cite{VidTIMIT} & FS\_GAN~\cite{FaceSwap-GAN} & \checkmark & \checkmark \\
    % \midrule
    Face Forensics++~\cite{faceforensics++} & Jan 2019 & 1,000 & 4,000 & 5,000 & YouTube & FS~\cite{FaceSwap}, DF & \checkmark & $\times$ \\
    % \midrule
    DFD~\cite{DFD} & Sep 2019 & 361 & 3,070 & 3,431 & YouTube & DF & \checkmark & $\times$ \\
    % \midrule
    CelebDF~\cite{Celeb-DF} & Nov 2019 & 408 & 795 & 1,203 & YouTube & DF & \checkmark & $\times$ \\
    % \midrule
    \rowcolor{blue!20} DFDC~\cite{DFDC} & Oct 2019 & 19,154 & 99,992 & 119,146 & Actors & Unknown & \checkmark & \checkmark \\
    % \midrule
    Deeper Forensics 1.0~\cite{deeperforensics} & Jan 2020 & 50,000 & 10,000 & 60,000 & Actors & DF & \checkmark & -- \\
    \bottomrule
    \end{tabular}
    }
    \vspace{-10pt}
\end{table*}

%% file: sections/02-relatedwork.tex
\section{Related Work}
\label{sec:relatedwork}
In this section we summarise prior work done in the domain. We first look into available literature in multimedia forensics in Section~\ref{subsec:forensics}. We elaborate on prior work in unimodal deepfake detection methods in Section~\ref{subsec:unimodal}. In Section~\ref{subsec:multimodal}, we discuss the multimodal approaches for deepfake detection. We summarize the deepfake video datasets in Section~\ref{subsec:datasets}. We give references from affective computing literature motivating the use of affective cues from modalities in our method for deepfake detection in Section~\ref{subsec:psychology}. 
%%%%%%%%%%%%%%%%%%%%%%%%%%%%%%%%%%%%%%%%%%%%%%%%%%%%%%%%%%%%%%%%%%%%%%
\subsection{Multimedia Forensics}
\label{subsec:forensics}

Media forensics deals with the problem of authenticating the source of digital media such as images, videos, speech, and text in order to identify forgeries and other malicious intents~\cite{for1}. Classical computer vision techniques~\cite{for1,for4} have sufficed to tackle media manually manipulated by humans. However recently, malicious actors have begun manipulating media using deep learning, \textit{e.g.} deepfakes~\cite{DeepFakesMalicious}, that use artificial intelligence to present false media as realistic in highly sophisticated and convincing ways. Classical computer vision techniques are unsuccessful when identifying false media corrupted using deep learning~\cite{for7}. Therefore, there has been a growing interest in developing deep learning-based methods for multimedia forensics~\cite{for3,for5,for6}. Ther has also been prior interest in exploring affect for detection deception~\cite{liar}. Our proposed method falls in this category and is complementary to other deep learning-based approaches.

% [To Add (1 line)] What is multimedia forensics? Stress that the area needs to evolve with the availability of media editing tools, altering images and videos has become widespread. Along with with ubiquitous social networks, this allows for the spread of fake news, hence the need of updated multimedia forensic techniques. [To Add (2-3 papers)] Some recent attempts in the area of mutimedia forensics.
%%%%%%%%%%%%%%%%%%%%%%%%%%%%%%%%%%%%%%%%%%%%%%%%%%%%%%%%%%%%%%%%%%%%%%
\subsection{Unimodal DeepFake Detection Methods}
\label{subsec:unimodal}
Most prior work in deepfake detection decompose videos into frames and explore visual artifacts across frames. For instance, Li et al.~\cite{FWA} propose a Deep Neural Network~(DNN) to detect fake videos based on artifacts observed during the face warping step of the generation algorithms. Similarly, Yang et al.~\cite{UADFV} look at inconsistencies in the head poses in the synthesized videos and Matern et al.~\cite{VA-MLP} capture artifacts in the eyes, teeth and facial contours of the generated faces. Prior works have also experimented with a variety of network architectures. For instance, Nguyen et al.~\cite{Capsule} explore capsule structures, Rossler et al.~\cite{faceforensics++} use the XceptionNet, and Zhou et al.~\cite{two-stream} use a two-stream Convolutional Neural Network~(CNN) to achieve SOTA in general-purpose image forgery detection. Previous researchers have also observed and exploited the fact that temporal coherence is not enforced effectively in the synthesis process of deepfakes. For instance, Sabir et al.~\cite{temporal1} leveraged the use of spatio-temporal features of video streams to detect deepfakes. Likewise, Guera and Delp et al. ~\cite{temporal2} highlight that deepfake videos contain intra-frame consistencies and hence use a CNN with a Long Short Term Memory~(LSTM) to detect deepfake videos.
%%%%%%%%%%%%%%%%%%%%%%%%%%%%%%%%%%%%%%%%%%%%%%%%%%%%%%%%%%%%%%%%%%%%%%
%%%%%%%%%%%%%%%%%%%%%%%%%%%%%%%%%%%%%%%%%%%%%%%%%%%%%%%%%%%%%%%%%%%%%%
\subsection{Multimodal DeepFake Detection Methods}
\label{subsec:multimodal}
While unimodal DeepFake Detection methods~(discussed in Section~\ref{subsec:unimodal}) have focused only on the facial features of the subject, there has not been much focus on using the multiple modalities that are part of the same video. Jeo and Bang et al.~\cite{multimodal1} propose FakeTalkerDetect, which is a Siamese-based network to detect the fake videos generated from the neural talking head models. They perform a classification based on distance. However, the two inputs to their Siamese network are a real and fake video. Korshunov et al.~\cite{multimodal2} analyze the lip-syncing inconsistencies using two channels, the audio and visual of moving lips. Krishnamurthy et al.~\cite{multimodal3} investigated the problem of detecting deception in real-life videos, which is very different from deepfake detection. They use an MLP-based classifier combining video, audio, and text with Micro-Expression features. Our approach to exploiting the mismatch between two modalities is quite different and complimentary to these methods.
\subsection{DeepFake Video Datasets}
\label{subsec:datasets}
The problem of deepfake detection has increased considerable attention, and this research has been stimulated with many datasets. We summarize and analyze $7$ benchmark deepfake video detection datasets in Table~\ref{tab:datasets}. Furthermore, the newer datasets~(DFDC~\cite{DFDC} and Deeper Forensics-1.0~\cite{deeperforensics})are larger and do not disclose details of the AI model used to synthesize the fake videos from the real videos.  Also, DFDC is the only dataset that contains a mix of videos with manipulated faces, audio, or both. All the other datasets contain only manipulated faces. Furthermore, only DFDC and DF-TIMIT~\cite{DeepFake-TIMIT} contain both audio and video, allowing us to analyze both modalities.
%%%%%%%%%%%%%%%%%%%%%%%%%%%%%%%%%%%%%%%%%%%%%%%%%%%%%%%%%%%%%%%%%%%%%%
\subsection{Affective Computing}
\label{subsec:psychology}
Understanding the perceived emotions of individuals using verbal and non-verbal cues is an important problem in both AI and psychology, especially when self-reported emotions are absent to be able to infer the actual emotions of the subjects~\cite{quigley2014inducing}. There is vast literature in inference of perceived emotions from a single modality or a combination of multiple modalities like facial expressions~\cite{face1,face2}, speech/audio signals~\cite{speech1}, body pose~\cite{body}, walking styles~\cite{step} and physiological features~\cite{physiological1}. There are also works exploring correlation in the affective features obtained from these various modalities.
Shan et al.~\cite{psych1} state that even if two modalities representing the same emotion vary in terms of appearance, the features detected are similar and should be correlated. Hence, if projected to a common space, they are compatible and can be fused to make inferences. Zhu. et al.~\cite{neural} explore the relationship between visual and auditory human modalities. Based on the neuroscience literature, they suggest that the visual and auditory signals are coded together in small populations of neurons within a particular part of the brain. \cite{sync1,sync2,sync3} explored the correlation of lip movements with speech. Studies concluded that our understanding of the speech modality is greatly aided by the sight of the lip and facial movements. Subsequently, such correlation among modalities has been explored extensively to perform multimodal emotion recognition~\cite{psych2,psych3,psych5}. These works have suggested and shown correlations between affect features obtained from the individual modalities~(face, speech, eyes, gestures). For instance, Mittal et al.~\cite{m3er} propose a multimodal perceived emotion perception network, where they use the correlation among modalities to differentiate between effectual and ineffectual modality features. Our approach is motivated by these developments in psychology research.
%%%%%%%%%%%%%%%%%%%%%%%%%%%%%%%%%%%%%%%%%%%%%%%%%%%%%%%%%%%%%%%%%%%%%%
% \subsection{Siamese Network Architecture and Triplet Loss}
% \label{subsec:siamese}
% The Siamese network~\cite{siamese1} architecture consists of two neural networks that share the same weights and are trained together. Each network typically takes in a different pattern (\textit{e.g.}, two views of an image, two tones of a speech), and the end output is a value representing the similarity between those two patterns. The overall network is trained using variants of the triplet loss or the contrastive loss, which are designed to maximize the distance between features learned from dissimilar patterns and minimize the distance between features learned from similar patterns. With this training objective, Siamese network-based architectures have been extensively used in applications such as face recognition~\cite{siamese2}, face verification~\cite{siamese3}, speaker identification~\cite{siamese4}, and even fake sample generation~\cite{mbgan}. In our work, we develop a Siamese network-based architecture and a variant of the triplet loss to maximally separate features learned from real and fake videos.
%%%%%%%%%%%%%%%%%%%%%%%%%%%%%%%%%%%%%%%%%%%%%%%%%%%%%%%%%%%%%%%%%%%%%%
\begin{figure*}[htp]
\centering
\subfloat[\small{\textbf{Training Routine: } \textit{(left)} We extract facial and speech features from the raw videos (each subject has a real and fake video pair) using OpenFace and pyAudioAnalysis, respectively. \textit{(right)} The extracted features are passed to the training network that consists of two modality embedding networks and two perceived emotion embedding networks. \label{fig:training}}]{%
  \includegraphics[width=0.7\textwidth]{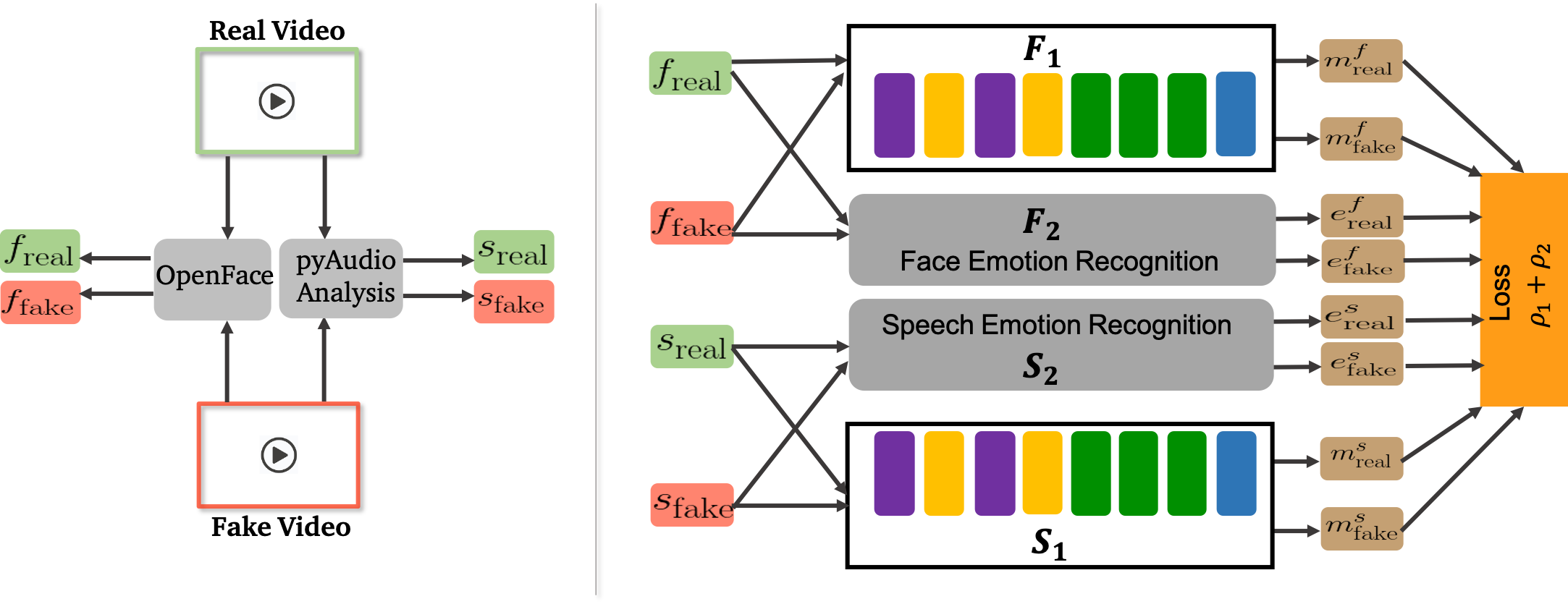}%
}\hfil
\subfloat[\small{\textbf{Testing Routine: } At runtime, given an input video, our network predicts the label (real or fake).\label{fig:testing}}]{%
\includegraphics[width=0.7\textwidth]{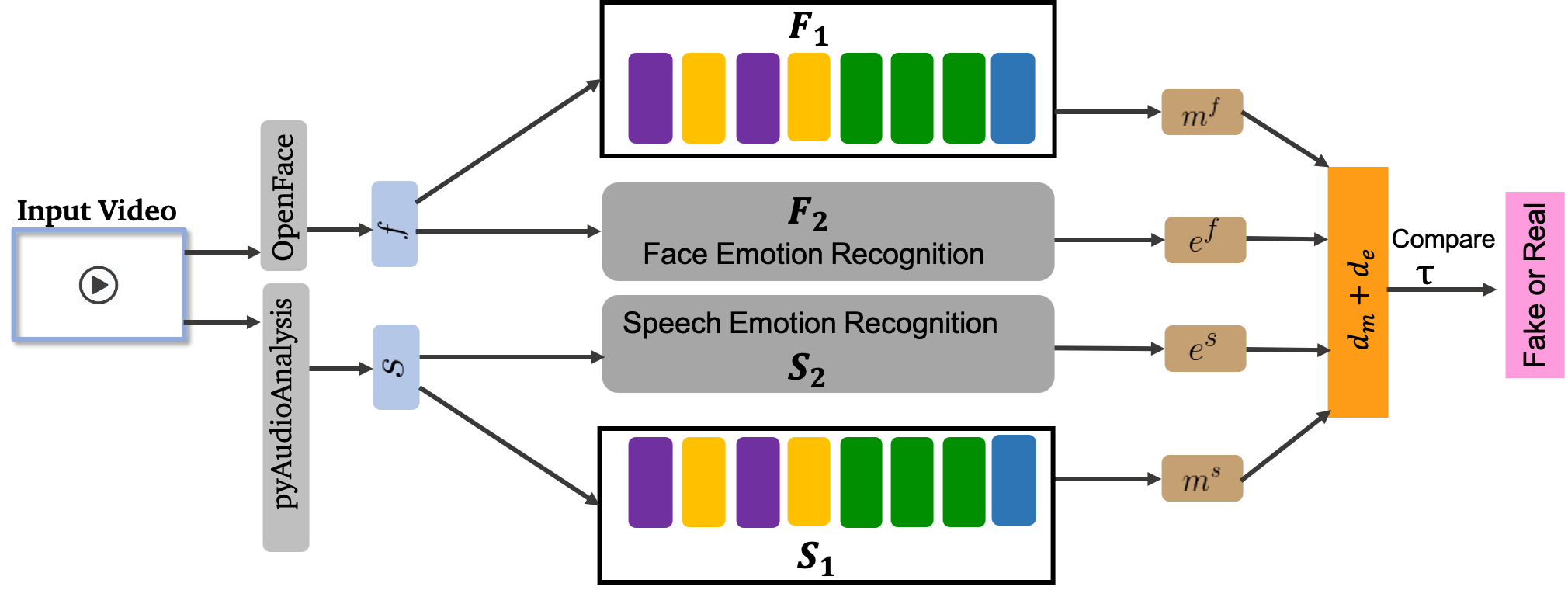}%
}
\caption{\small{\textbf{Overview Diagram: }We present an overview diagram for both the training and testing routines of our model. The networks consist of 2D convolutional layers~(purple), max-pooling layers~(yellow), fully-connected layers~(green), and normalization layers~(blue). $F_1$ and $S_1$ are modality embedding networks and $F_2$ and $S_2$ are perceived emotion embedding networks for face and speech, respectively.}}
\label{fig:overview}
\end{figure*}
%%%%%%%%%%%%%%%%%%%%%%%%%%%%%%%%%%%%%%%%%%%%%%%%%%%%%%%%%%%%%%%%%%%%%%

%% file: sections/03-approach.tex
\section{Our Approach}
\label{sec:approach}
%%%%%%%%%%%%%%%%%%%%%%%%%%%%%%%%%%%%%%%%%%%%%%%%%%%%%%%%%%%%%%%%%%%%%%
In this section, we present our multimodal approach to detecting deepfake videos. We briefly describe the problem statement and give an overview of our approach in Section~\ref{subsec:problemstatement}. We also elaborate on how our approach is similar to a Siamese Network architecture. We elaborate on the modality embeddings and the perceived emotion embedding, the two main components in Section~\ref{subsec:embedding} and Section~\ref{subsec:emotion}, respectively. We explain the similarity score and modified triplet losses used for training the network in Section~\ref{subsec:training}, and finally, in Section~\ref{subsec:testing}, we explain how they are used to classify between real and fake videos. We list all notations used throughout the paper in Table~\ref{tab: notation}.

%%%%%%%%%%%%%%%%%%%%%%%%%%%%%%%%%%%%%%%%%%%%%%%%%%%%%%%%%%%%%%%%%%%
\input{tables/notation}

%%%%%%%%%%%%%%%%%%%%%%%%%%%%%%%%%%%%%%%%%%%%%%%%%%%%%%%%%%%%%%%%%%%

%%%%%%%%%%%%%%%%%%%%%%%%%%%%%%%%%%%%%%%%%%%%%%%%%%%%%%%%%%%%%%%%%%%%%%%
\subsection{Problem Statement and Overview}
\label{subsec:problemstatement}
Given an input video with audio-visual modalities present, our goal is to detect if it is a deepfake video. Overviews of our training and testing routines are given in Figure~\ref{fig:training} and Figure~\ref{fig:testing}, respectively. During training, we select one ``real'' and one ``fake'' video containing the same subject. We extract the visual face as well as the speech features, $\fr$ and $\sr$, respectively, from the real input video. In similar fashion, we extract the face and speech features~(using OpenFace~\cite{openface} and pyAudioAnalysis~\cite{pyaudioanalysis}), $\ff$ and $\spf$, respectively, from the fake video. More details about the feature extraction from the raw videos have been presented in Section~\ref{subsec:featureextraction}. The extracted features, $\fr, \sr, \ff, \spf$, form the inputs to the networks~(\texttt{$F_1$}, \texttt{$F_2$}, \texttt{$S_1$}, and \texttt{$S_2$}), respectively. We train these networks using a combination of two triplet loss functions designed using the similarity scores, denoted by $\rho_1$ and $\rho_2$. $\rho_1$ represents similarity among the facial and speech modalities, and $\rho_2$ is the similarity between the affect cues~(specifically, perceived emotion) from the modalities of both real and fake videos.

Our training method is similar to a Siamese network because we also use the same weights of the network~($F_1, F_2, S_1, S_2$) to operate on two different inputs, one real video and the other a fake video of the same subject. Unlike regular classification-based neural networks, which perform classification and propagate that loss back, we instead use similarity-based metrics for distinguishing the real and fake videos. We model this similarity between these modalities using Triplet loss~(explained elaborately in Section~\ref{subsec:testing}). 

During testing, we are given a single input video, from which we extract the face and speech feature vectors, $f$ and $s$, respectively. We pass $f$ into \texttt{$F_1$} and \texttt{$F_2$}, and pass $s$ into \texttt{$S_1$} and \texttt{$S_2$}, where \texttt{$F_1$}, \texttt{$F_2$}, \texttt{$S_1$}, and \texttt{$S_2$} are used to compute distance metrics, $dist_1$ and $dist_2$. We use a threshold $\tau$, learned during training, to classify the video as real or fake. 

% \vspace{-20pt}
\subsection{$F_1$ and $S_1$: Video/Audio Modality Embeddings}
\label{subsec:embedding}
%%%%%%%%%%%%%%%%%%%%%%%%%%%%%%%%%%%%%%%%%%%%%%%%%%%%%%%%%%%%%%%%%%%%%%
\texttt{$F_1$} and \texttt{$S_1$} are neural networks that we use to learn the unit-normalized embeddings for the face and speech modalities, respectively. In Figure~\ref{fig:overview}, we depict $F_1$ and $S_1$ in both training and testing routines. They are composed of $2$D convolutional layers~(purple), max-pooling layers~(yellow), and fully connected layers~(green). ReLU non-linearity is used between all layers. The last layer is a unit-normalization layer~(blue). For both face and speech modalities, $F_1$ and $S_1$ return $250$-dimensional unit-normalized embeddings.

% \textbf{Training}:
The training is performed using the following equations:
\begin{equation}
\begin{split}
     m^f_\textrm{real} = F_1(\fr), \ &m^f_\textrm{fake} = F_1(\ff), \\
     m^s_\textrm{real} = S_1(\sr), \ &m^s_\textrm{fake} = S_1(\spf)
\end{split}
\label{eq:embedding_train}
\end{equation}

And the testing is done using the equations:

% \textbf{Testing}:
\begin{equation}
m^f =  F_1(f), \ m^s =  S_1(s)
\label{eq:embedding_test}
\end{equation}
%%%%%%%%%%%%%%%%%%%%%%%%%%%%%%%%%%%%%%%%%%%%%%%%%%%%%%%%%%%%%%%%%%%
\subsection{$F_2$ and $S_2$: Video/Audio Perceived Emotion Embedding}
\label{subsec:emotion}
\texttt{$F_2$} and \texttt{$S_2$} are neural networks that we use to learn the unit-normalized affect embeddings for the face and speech modalities, respectively. \texttt{$F_2$} and \texttt{$S_2$} are based on the Memory Fusion Network (MFN)~\cite{mfn}, which is reported to have SOTA performance on both emotion recognition from multiple views or modalities like face and speech. MFN is based on a recurrent neural network architecture with three main components: a system of LSTMs, a Memory Attention Network, and a Gated Memory component. The system of LSTMs takes in different views of the input data. In our case, we adopt the trained single-view version of the MFN, where the face and speech are treated as separate views, i.e. \texttt{$F_2$} takes in the video~(view only) and \texttt{$S_2$} takes in the audio~(view only). We pre-trained the \texttt{$F_2$} MFN with video and the \texttt{$S_2$} MFN with audio from CMU-MOSEI dataset~\cite{cmu-mosei}. The CMU-MOSEI dataset describes the perceived emotion space with 6 discrete emotions following the Ekman model~\cite{ekman_model}: happy, sad, angry, fearful, surprise, and disgust, and a ``neutral'' emotion to denote the absence of any of these emotions. For face and speech modalities in our network, we use $250$-dimensional unit-normalized features constructed from the cross-view patterns learned by $F_2$ and $S_2$ respectively.

% \begin{equation}
% \begin{split}
%     &\textrm{\textbf{Training}: }  e^f_\textrm{real} = F_2(\fr), e^f_\textrm{fake} = F_2(\ff), e^s_\textrm{real} = S_2(\sr), e^s_\textrm{fake} = S_2(\spf). \\
% &\textrm{\textbf{Testing}: }e^f =  F_2(f), \ e^s =  S_2(s).
% \end{split}
% \label{eq:emotions}
% \end{equation}
The training is performed using the following equations:

% \textbf{Training}:
\begin{equation}
\begin{split}
     e^f_\textrm{real} = F_2(\fr), e^f_\textrm{fake} = F_2(\ff), \\
     e^s_\textrm{real} = S_2(\sr), e^s_\textrm{fake} = S_2(\spf).
\end{split}
\label{eq:emotion_train}
\end{equation}
% \textbf{Testing}:
And the testing is done using the equations:
\begin{equation}
e^f =  F_2(f), \ e^s =  S_2(s).
\label{eq:emotion_test}
\end{equation}

\vspace{-10pt}
%%%%%%%%%%%%%%%%%%%%%%%%%%%%%%%%%%%%%%%%%%%%%%%%%%%%%%%%%%%%%%%%%%%%%%
\subsection{Training Routine}
\label{subsec:training}
At training time, we use a fake and a real video with the same subject as the input. After, passing extracted features from raw videos~($\fr, \ff, \sr, \spf$) through $F_1$, $F_2$, $S_1$ and $S_2$, we obtain the unit-normalised modality and perceived emotion embeddings as described in Eqs.~\ref{eq:embedding_train}-\ref{eq:emotion_test}.

Considering a input real and fake video, we first compare $\fr$ with  $\ff$, and $\sr$ with $\spf$ to understand what modality was manipulated \textit{more} in the fake video. Considering, we identify the face modality to be manipulated more in the fake video, based on these embeddings we compute the first similarity between the real and fake speech and face embeddings as follows: 
\begin{equation}
\textrm{\textbf{Similarity Score 1}: }    L_1 = d(m^s_\textrm{real}, m^f_\textrm{real}) - d(m^s_\textrm{real}, m^f_\textrm{fake}),
\end{equation}
where $d$ denotes the Euclidean distance. 

In simpler terms, $L_1$ is computing the distance between two pairs, $d(m^s_\textrm{real}, m^f_\textrm{real})$ and $d(m^s_\textrm{real}, m^f_\textrm{fake})$. We expect $m^s_\textrm{real}, m^f_\textrm{real}$ to be closer to each other than $m^s_\textrm{real}, m^f_\textrm{fake}$ as it contains a fake face modality. Hence, we expect to maximize this difference. To use this correlation metric as a loss function to train our model, we formulate it using the notation of Triplet Loss
\begin{equation}
\textrm{\textbf{Similarity Loss 1}: }     \rho_1 = \max \big(L_1 + m_1, 0 \big),
\end{equation}
where $m_1$ is the margin used for convergence of training. 

If we had observed that speech is the more manipulated modality in the fake video, we would formulate $L_1$ as follows:
\[  L_1 = d(m^f_\textrm{real}, m^s_\textrm{real}) - d(m^f_\textrm{real}, m^s_\textrm{fake}). \] 

Similarly, we compute the second similarity as the difference in affective cues extracted from the modalities from both real and fake videos. We denote this as follows:

\begin{equation}
\textrm{\textbf{Similarity Score 2}: } L_2 = d(e^s_\textrm{real}, e^s_\textrm{fake}) - d(f^s_\textrm{real},e^f_\textrm{fake}). 
\end{equation}

As per prior psychology studies, we expect that similar un-manipulated modalities point towards similar affective cues. Hence, because the input here has a manipulated face modality, we expect $e^s_\textrm{real}, e^s_\textrm{fake}$ to be closer to each other than to $e^f_\textrm{real}, e^f_\textrm{fake}$. To use this as a loss function, we again formulate this using a Triplet loss. 
\begin{equation}
\textrm{\textbf{Similarity Loss 2}: }     \rho_2 = \max(L_2 + m_2,0),
\end{equation}
where $m_2$ is the margin. 

Again, if speech was the highly manipulated modality in the fake video, we would formulate $L_2$ as follows:
\[  L_2 = d(e^f_\textrm{real}, e^f_\textrm{fake}) - d(e^f_\textrm{real},e^s_\textrm{fake}). \]
We use both the similarity scores as the cumulative loss and propagate this back into the network. 
\begin{equation}
    Loss = \rho_1 + \rho_2
\end{equation}
%%%%%%%%%%%%%%%%%%%%%%%%%%%%%%%%%%%%%%%%%%%%%%%%%%%%%%%%%%%%%%%%%%%%%%
\subsection{Testing Routine}
\label{subsec:testing}
At test time, we only have a single input video that is to be labeled real or fake. After extracting the features, $f$ and $s$ from the raw videos, we perform a forward pass through $F_1$, $F_2$, $S_1$ and $S_2$, as depicted in Figure~\ref{fig:testing} to obtain modality and perceived emotion embeddings. 

To make an inference about real and fake, we compute the following two distance values:
\begin{equation}
    \begin{split}
         &\textrm{\textbf{Distance 1}: } d_m = d(m^f, m^s),\\
         &\textrm{\textbf{Distance 2}: } d_e = d(e^f, e^s).
    \end{split}
    \label{eq: distance_functions}
\end{equation}
%  \textrm{\textbf{Distance 1}: } d_m = d(m^f, m^s)
% \end{eq}
% \begin{eq}
% \textrm{\textbf{Distance 2}: } d_e = d(e^f, e^s)
% \end{eq}

To distinguish between real and fake, we compare $d_m$ and $d_e$ with a threshold, that is, $\tau$ empirically learned during training as follows:
\[  \textrm{If } d_m + d_e > \tau, \] we label the video as a fake video. 

\underline{Computation of $\tau$: }To compute $\tau$, we use the best-trained model and run it on the training set. We compute $d_m$ and $d_e$ for both real and fake videos of the train set. We average these values and find an equidistant number, which serves as a good threshold value. Based on our experiments, the computed value of $\tau$ was almost consistent and didn't vary much between datasets.

%% file: tables/notation.tex
\begin{table}
  \caption{\small{\textbf{Notation: } We highlight the notation and symbols used in the paper.}}
  \label{tab: notation}
  \centering
  \resizebox{\columnwidth}{!}{%
  \begin{tabular}{ll}
  \toprule
    \textbf{Symbol}  & \textbf{Description} \\
    \midrule
    \multirow{3}{*}{$x_y$} & $x \in \{ f,s \}$ denote face and speech features extracted from OpenFace and pyAudioAnalysis.\\
                           & $y \in \{ \textrm{real},\textrm{fake} \}$ indicate whether the feature $x$ is real or fake.\\
                           &E.g. $f_\textrm{real}$ denotes the \textit{face features} extracted from a \textit{real} video using OpenFace.\\
                           \midrule
    \multirow{4}{*}{$a^b_c$} & $a \in \{ e,m \}$ denote emotion embedding and modality embedding.\\
                             & $b \in \{ f,s \}$ denote face and speech cues.\\
                             & $c \in \{ \textrm{real},\textrm{fake} \}$ indicate whether the embedding $a$ is real or fake.\\
                             & E.g. $m^f_\textrm{real}$ denotes the \textit{face modality} embedding generated from a \textit{real} video.\\
                             \midrule
    $\rho_1$                 & Modality Embedding Similarity Loss (Used in Training)\\
    $\rho_2$                 & Emotion Embedding Similarity Loss (Used in Training)\\
                            %  &$\rho_m, \rho_e$ can be positive or negative depending on whether the \\
                             \midrule
    $d_m$                 & Face/Speech Modality Embedding Distance (Used in Testing)\\
    $d_e$                 & Face/Speech Emotion Embedding Distance (Used in Testing)\\
    \bottomrule
  \end{tabular}
   }
   \vspace{-10pt}
\end{table}

%% file: sections/04-implementation.tex
\section{Implementation and Evaluation}
\label{sec:implementation}
\subsection{Datasets}
\label{subsec:used-datasets}
We perform experiments on the DF-TIMIT~\cite{DeepFake-TIMIT} and DFDC~\cite{DFDC} datasets, as only these datasets contain modalities for face and speech features (\ref{tab:datasets}). We used the entire DF-TIMIT dataset and were able to use randomly sampled $18,000$ videos from DFDC dataset due to computational overhead. Both the datasets are split into training ($85$\%), and testing ($15$\%) sets. 

%%%%%%%%%%%%%%%%%%%%%%%%%%%%%%%%%%%%%%%%%%%%%%%%%%%%%%%%%%%%%%%%%%%%%%
\subsection{Training Parameters}
\label{subsec:hyperparameters}
On the DFDC Dataset, we trained our models with a batch size of $128$ for $500$ epochs. Due to the significantly smaller size of the DF-TIMIT dataset, we used a batch size of $32$ and trained it for $100$ epochs. We used Adam optimizer with a learning rate of $0.01$. All our results were generated on an NVIDIA GeForce GTX1080 Ti GPU.

%%%%%%%%%%%%%%%%%%%%%%%%%%%%%%%%%%%%%%%%%%%%%%%%%%%%%%%%%%%%%%%%%%%%%%
\subsection{Feature Extraction}
In our approach (See Figure~\ref{fig:overview}), we first extract the face and speech features from the real and fake input videos. We use existing SOTA methods for this purpose. In particular, we use OpenFace~\cite{openface} to extract $430$-dimensional facial features, including the $2$D landmarks positions, head pose orientation, and gaze features. To extract speech features, we use pyAudioAnalysis~\cite{pyaudioanalysis} to extract $13$ Mel Frequency Cepstral Coefficients (MFCC) speech features. Prior works~\cite{audio1,audio2,audio3} using audio or speech signals for various tasks like perceived emotion recognition, and speaker recognition use MFCC features to analyse audio signals.
\label{subsec:featureextraction}

%% file: sections/05-results.tex
\section{Results and Analysis}
\label{sec:results}
In this section, we elaborate on some quantitative and qualitative results of our methods.
%%%%%%%%%%%%%%%%%%%%%%%%%%%%%%%%%%%%%%%%%%%%%%%%%%%%%%%%%%%%%%%%%%%%%%
\subsection{Comparison with SOTA Methods}
\label{subsec:sota}
We report and compare per-video AUC Scores of our method against $9$ prior deepfake video detection methods on DF-TIMIT and DFDC. To ensure a fair evaluation, while the subset of DFDC the $9$ methods were trained and tested are unknown, we select $18,000$ samples randomly and report our numbers. Moreover, as per the nature of the approaches the prior $9$ methods report per-frame AUC scores. We have summarized these results in Table~\ref{tab:auc_scores}. The following are the prior methods used to compare the performance of our approach on the same datasets. 
\begin{enumerate}[noitemsep]
    \item \underline{Two-stream}~\cite{two-stream}: uses a two-stream CNN to achieve SOTA performance in image-forgery detection. They use standard CNN network architectures to train the model. 
    \item \underline{MesoNet}~\cite{Mesonet} is a CNN-based detection method that targets the microscopic properties of images. AUC scores are reported on two variants. 
    \item \underline{HeadPose}~\cite{UADFV} captures inconsistencies in headpose orientation across frames to detect deepfakes. 
    \item \underline{FWA}~\cite{FWA} uses a CNN to expose the face warping artifacts introduced by the resizing and interpolation operations. 
    \item \underline{VA}~\cite{VA-MLP} focuses on capturing visual artifacts in the eyes, teeth and facial contours of synthesized faces. Results have been reported on two standard variants of this method.
    \item \underline{Xception}~\cite{faceforensics++} is a baseline model trained on the FaceForensics++ dataset based on the XceptionNet model. AUC scores have been reported on three variants of the network. 
    \item \underline{Multi-task}~\cite{MultiTask} uses a CNN to simultaneously detect manipulated images and segment manipulated areas as a multi-task learning problem.
    \item \underline{Capsule}~\cite{Capsule} uses capsule structures based on a standard DNN. 
    \item \underline{DSP-FWA} is an improved version of FWA~\cite{FWA} with a spatial pyramid pooling module to better handle the variations in resolutions of the original target faces.
\end{enumerate}
\begin{figure*}
    \centering
    \includegraphics[width=0.85\textwidth]{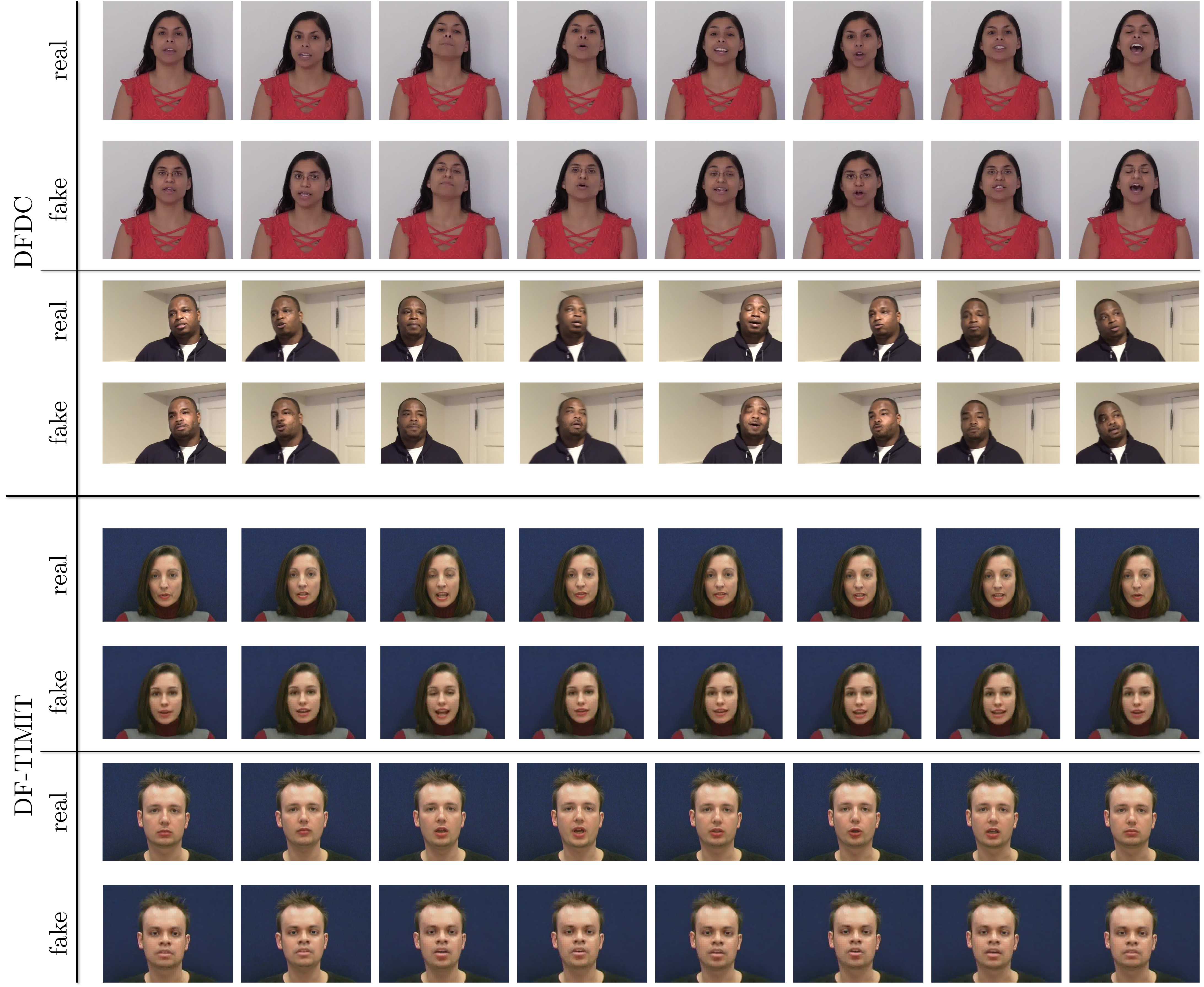}
    \caption{\small{\textbf{Qualitative Results:} We show results of our model on the DFDC and DF-TIMIT datasets. Our model uses the subjects' audio-visual modalities as well as their perceived emotions to distinguish between real and deepfake videos. The perceived emotions from the speech and facial cues in fake videos are different; however in the case of real videos, the perceived emotions from both modalities are the same.}}
    \label{fig:qualitative}
    \vspace{-10pt}
\end{figure*}
\input{tables/results}
While we outperform on the DFDC dataset, we have comparable values for the DF-TIMIT dataset. We believe this is because all $640$ videos in the DF-TIMIT dataset are face-centered with no body pose. In DFDC, the videos are collected with full-body poses, with the face taking less than $50\%$ of the pixels in each frame. The FWA and DSP-FWA methods identify deepfakes by detecting artifacts caused by affine warping of manipulated faces to match the configuration of the source’s face. This is especially useful for the face-centered DF-TIMIT dataset than the DFDC dataset. 

%%%%%%%%%%%%%%%%%%%%%%%%%%%%%%%%%%%%%%%%%%%%%%%%%%%%%%%%%%%%%%%%%%%%%%%%%%%%%
% \begin{figure*}[htp]
% \centering
% \subfloat[\small{\textbf{Correct Classification: }Our model correctly classified this popular deepfake video as fake.}]{%
%   \includegraphics[width=0.85\textwidth]{images/trump_8_frames.png}%
% }\hfill
% \subfloat[\small{\textbf{Misclassification:} Our model was not able to detect this deepfake video as fake because the network was not able to detect any perceived emotion from the subject.  }]{%
% \includegraphics[width=0.85\textwidth]{images/obama_8_frames.png}%
% }
% \caption{\textbf{In the Wild Deepfake Videos:} Our model succeeds in the wild. We collect several popular deepfake videos from online social media and our model achieves reasonably good results. We will present additional results in the supplementary video. }
%     \label{fig:wild}
%     \vspace{-10pt}
% \end{figure*}
% \vspace{-10pt}
%%%%%%%%%%%%%%%%%%%%%%%%%%%%%%%%%%%%%%%%%%%%%%%%%%%%%%%%%%%%%%%%%%%%%%
\subsection{Qualitative Results}
\label{subsec:qual}
We show some selected frames of videos from both the datasets in Figure~\ref{fig:qualitative} along with the labels~(real/fake). For the qualitative results shown for DFDC, the real video predicted a ``neutral'' perceived emotion label for both speech and face modality, whereas in the fake video the face predicted ``surprise'' and speech predicted ``neutral''. This result is indeed interpretable because the fake video was generated by manipulating only the face modality and not the speech modality. We see a similar perceived emotion label mismatch for the DF-TIMIT sample as well.

%%%%%%%%%%%%%%%%%%%%%%%%%%%%%%%%%%%%%%%%%%%%%%%%%%%%%%%%%%%%%%%%%%%%%%
\subsection{Interpreting the Correlations}
\label{subsec:interpretability}
\begin{figure*}[t]
\centering
\subfloat[\small{\textbf{Modality Embedding Distances: }We plot the percentage of subject videos versus the distance between the face and speech modality embeddings. The figure shows that the distribution of real videos (blue curve) is centered around a lower modality embedding distance ($0.2$). In contrast, the fake videos (orange curve) are distributed around a higher distance center ($0.5$). \textbf{Conclusion:} We show that audio-visual modalities are more similar in real videos as compared to fake videos. }]{%
  \includegraphics[width=0.95\columnwidth]{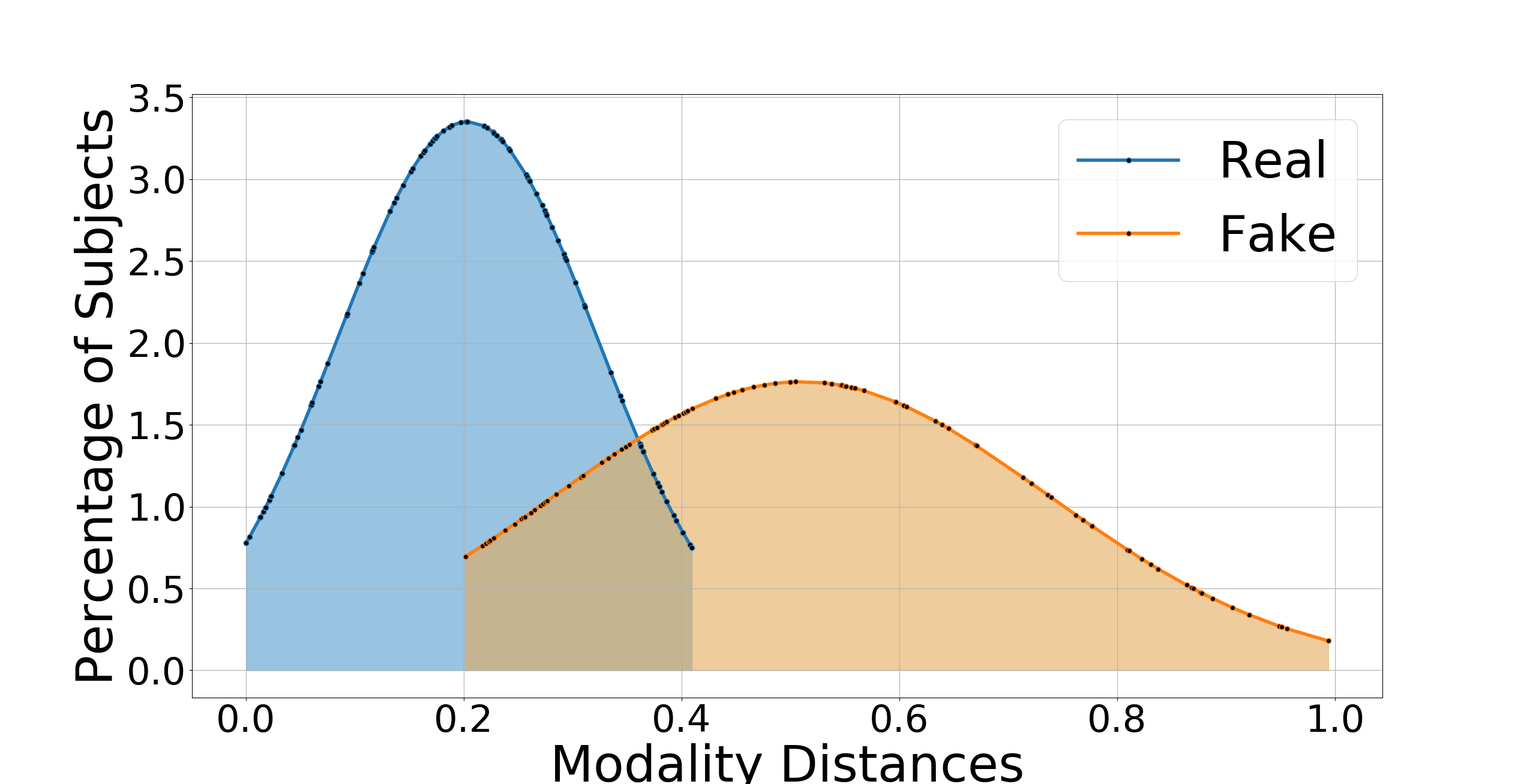}%
}\hfil
\subfloat[\small{\textbf{Perceived Emotion Embedding in Real and Fake Videos: }The blue and orange bars represent the total number of videos where the perceived emotion labels, obtained from the face and speech modalities, do \textit{not} match, and match, respectively. Of the total $15,438$ fake videos, $73.2\%$ videos were found to contain a mismatch between perceived emotion labels and for real videos this was only $24\%$.~\textbf{Conclusion: }We show that perceived emotions of subjects, from multiple modalities, are strongly similar in real videos, and often mismatched in fake videos.}]{%
  \includegraphics[width=0.95\columnwidth]{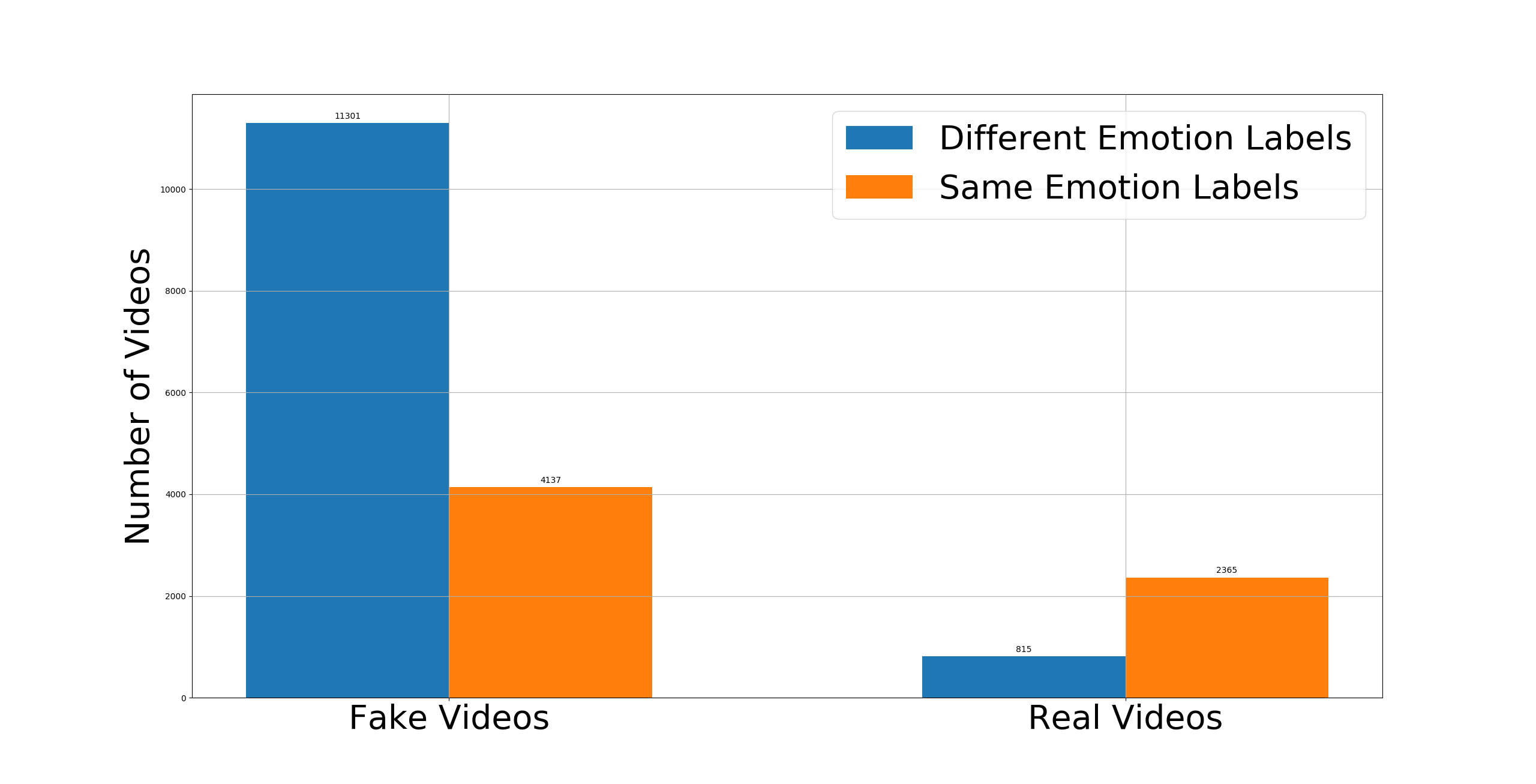}%
}
\caption{\small{\textbf{Embedding Interpretation: }We provide an intuitive interpretation of the learned embeddings from $F_1, S_1, F_2, S_2$ with visualizations. These results back our hypothesis of perceived emotions being highly correlated in real videos as compared to fake videos.}}
\label{fig:interpret}
\end{figure*}

To better understand the learned embeddings, we plot the distance between the unit-normalized face and speech embeddings learned from $F_1$ and $S_1$ on $1,000$ randomly chosen points from the DFDC train set in Figure~\ref{fig:interpret}(a). We plot $d(m^s_\textrm{real}, m^f_\textrm{real})$ in blue and $d(m^s_\textrm{fake}, m^f_\textrm{fake})$ in orange. It is interesting to see that the peak or the majority of the subjects from real videos have a smaller separation, $0.2$ between their embeddings as opposed to the fake videos~($0.5$). We also plot the number of videos, both fake and real, with a mismatch of perceived emotion labels extracted using $F_2$ and $S_2$ in Figure~\ref{fig:interpret}(b). Of a total of $15,438$ fake videos, $11,301$ showed a mismatch in the labels extracted from face and speech modalities. Similarly, out of $3,180$ real videos, $815$ also showed a label mismatch.

%%%%%%%%%%%%%%%%%%%%%%%%%%%%%%%%%%%%%%%%%%%%%%%%%%%%%%%%%%%%%%%%%%%%%%
%%%%%%%%%%%%%%%%%%%%%%%%%%%%%%%%%%%%%%%%%%%%%%%%%%%%%%%%%%%%%%%%%%%%%%
\input{tables/ablation}
%%%%%%%%%%%%%%%%%%%%%%%%%%%%%%%%%%%%%%%%%%%%%%%%%%%%%%%%%%%%%%%%%%%%%%
\begin{figure*}[t]
    \centering
    \includegraphics[width=0.85\textwidth]{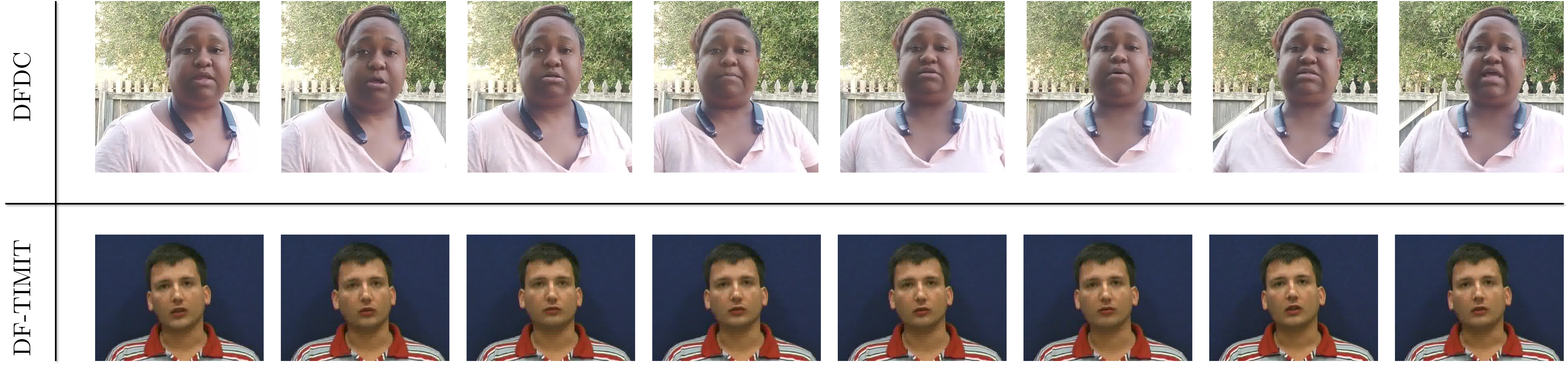}
    \caption{\small{\textbf{Misclassification Results: }We show one sample each from DFDC and DF-TIMIT where our model predicted the two fake videos as real due to incorrect perceived emotion embeddings.}}
    \label{fig:failure}
    \vspace{-15pt}
\end{figure*}
%%%%%%%%%%%%%%%%%%%%%%%%%%%%%%%%%%%%%%%%%%%%%%%%%%%%%%%%%%%%%%%%%%%%%%
\begin{figure*}[t]
    \centering
    \includegraphics[width=0.85\textwidth]{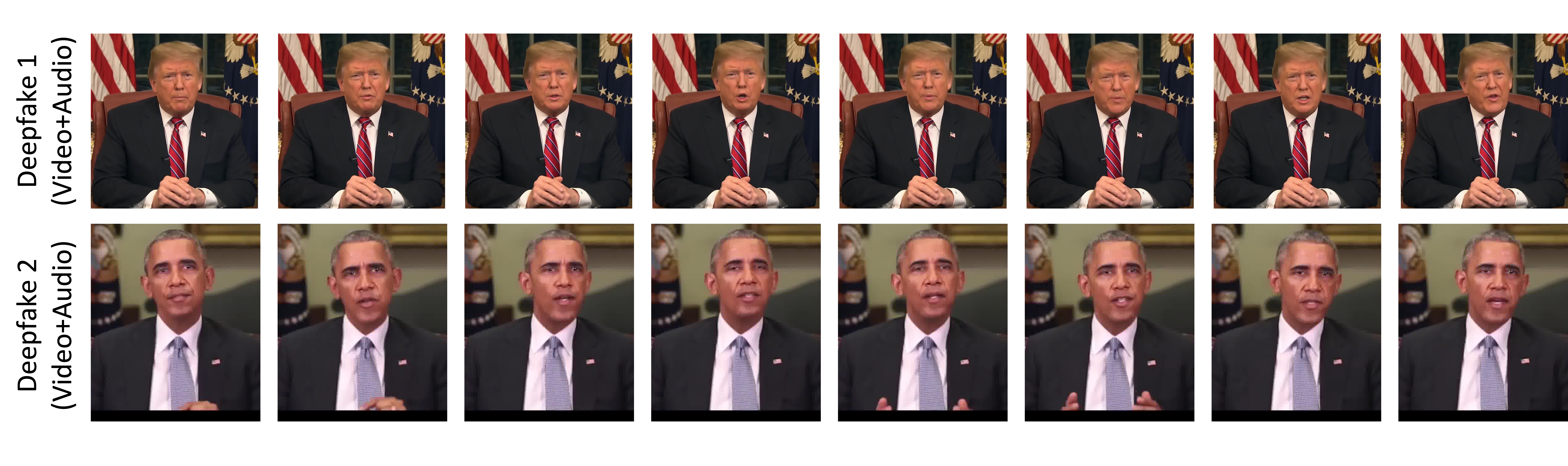}
    \caption{\small{\textbf{Results on In-The-Wild Videos: }Our model succeeds in the wild. We collect several popular deepfake videos from online social media and our model achieves reasonably good results.}}
    \label{fig:wild}
    \vspace{-15pt}
\end{figure*}
%%%%%%%%%%%%%%%%%%%%%%%%%%%%%%%%%%%%%%%%%%%%%%%%%%%%%%%%%%%%%%%%%%%%%%
\vspace{-5pt}
\subsection{Ablation Experiments}
\label{subsec:ablation}
As explained in Section~\ref{subsec:testing}, we use two distances, based on the modality embedding similarities and perceived emotion embedding similarities, to detects fake videos. To understand and motivate the contribution of each similarity, we perform an ablation study where we run the model using only one correlation for training. We have summarized the results of the ablation experiments in Table~\ref{tab:ablation}. The modality embedding similarity helps to achieve better AUC scores than the perceived emotion embedding similarity.

%%%%%%%%%%%%%%%%%%%%%%%%%%%%%%%%%%%%%%%%%%%%%%%%%%%%%%%%%%%%%%%%%%%%%%

\subsection{Failure Cases}
\label{subsec:failure}
Our approach models the correlation between two modalities and the associated affective cues to distinguish between real and fake modalities. However, there are multiple instances where the deepfake videos do not contain such a mismatch in terms of perceived emotional classification based on different modalities. This is also because humans express perceived emotions differently. As a result, our model fails to classify such videos as fake. Similarly, both face and speech are modalities that are easy to fake. As a result, it is possible that our method also classifies a real video as a fake video due to this mismatch. In Figure~\ref{fig:failure}, we show one such video from both the datasets, where our model failed.

%%%%%%%%%%%%%%%%%%%%%%%%%%%%%%%%%%%%%%%%%%%%%%%%%%%%%%%%%%%%%%%%%%%%%%
\subsection{Results on Videos in the Wild}
\label{subsec:failure}
We tested the performance of our model on two such deepfake videos obtained from an online social platform~\cite{obama,trump}. Some frames from this video have been shown in Figure~\ref{fig:wild}. While the model successfully classified the first video as a deepfake, it could not for the second deepfake video. 

%% file: tables/results.tex
\begin{table}[th]
    \centering
    \caption{\small{\textbf{AUC Scores.} Blue denotes best and green denotes second-best. Our model improves the SOTA by approximately 9\% on the DFDC dataset and achieves accuracy similar to the SOTA on the DF-TIMIT dataset.}}
    \label{tab:auc_scores}
    \resizebox{.9\columnwidth}{!}{
    \begin{tabular}{clccc}
    \toprule
    \multirow{ 3}{*}{\textbf{S.No.}} & \multirow{ 3}{*}{\textbf{Methods}} & \multicolumn{3}{c}{\textbf{Datasets}} \\
    \cmidrule{3-5}
    & & \multicolumn{2}{c}{\textbf{DF-TIMIT}~\cite{DeepFake-TIMIT}} & \textbf{DFDC}~\cite{DFDC} \\
    \cmidrule{3-4}
    & & \textbf{LQ} & \textbf{HQ} & \\
    \midrule
    1 & Capsule~\cite{Capsule} & 78.4 & 74.4 & 53.3 \\
    2 & Multi-task~\cite{MultiTask} & 62.2 & 55.3 & 53.6 \\
    3 & HeadPose~\cite{UADFV} & 55.1 & 53.2 & 55.9 \\
    4 & Two-stream~\cite{two-stream} & 83.5 & 73.5 & 61.4 \\
    \midrule
    \multirow{ 2}{*}{5} & VA-MLP~\cite{VA-MLP} & 61.4 & 62.1 & 61.9 \\
    & VA-LogReg & 77.0 & 77.3 & 66.2 \\
    \midrule
    \multirow{ 2}{*}{6}  & MesoInception4 & 80.4 & 62.7 & 73.2 \\
    & Meso4~\cite{Mesonet} & 87.8 & 68.4 & 75.3 \\
    \midrule
    % \midrule
    \multirow{ 3}{*}{7} & Xception-raw~\cite{faceforensics++} & 56.7 & 54.0 & 49.9 \\
    & Xception-c40 & 75.8 & 70.5 & 69.7 \\
    & Xception-c23 & 95.9 & 94.4 & 72.2 \\
    \midrule
    \multirow{ 2}{*}{8} & FWA~\cite{FWA} &  \cellcolor{blue!20}\textbf{99.9} & 93.2 & 72.7 \\
    & DSP-FWA & \cellcolor{blue!20}\textbf{99.9} &  \cellcolor{blue!20}\textbf{99.7} & \cellcolor{green!20}75.5 \\ 
    \midrule
    & \textbf{Our Method} & \cellcolor{green!20}96.3 & \cellcolor{green!20}94.9 &  \cellcolor{blue!20}\textbf{84.4}\\
    \bottomrule
    \end{tabular}
    }
    \vspace{-10pt}
    
\end{table}

%% file: tables/ablation.tex
\begin{table}[t]
    \centering
    \caption{\small{\textbf{Ablation Experiments.} To motivate our model, we perform ablation studies where we remove one correlation at a time for training and report the AUC scores.}}
    \label{tab:ablation}
    \resizebox{\columnwidth}{!}{%
    \begin{tabular}{lccc}
    \toprule
    \multirow{3}{*}{\textbf{Methods}} & \multicolumn{3}{c}{\textbf{Datasets}} \\
    \cline{2-4}
    & \multicolumn{2}{c}{\textbf{DF-TIMIT} \Tstrut~\cite{DeepFake-TIMIT}} & \textbf{DFDC}~\cite{DFDC} \\
    \cline{2-3}
    & \textbf{LQ} & \textbf{HQ} & \\
    \midrule
    Our Method w/o Modality Similarity ($\rho_1$) & 92.5 & 91.7 & 78.3 \\
    % \midrule
    Our Method w/o Emotion Similarity ($\rho_2$) & 94.8 & 93.6 & 82.8\\
    \midrule
  \textbf{Our Method }& \textbf{96.3} & \textbf{94.9} & \textbf{84.4}\\
    \bottomrule
    \end{tabular}
    }
% \vspace{-10pt}
\end{table}

%% file: sections/06-conclusion.tex
\vspace{-10pt}
\section{Conclusion, Limitations and Future Work}
\label{sec:conclusion}
We present a learning-based method for detecting fake videos. We use the similarity between audio-visual modalities and the similarity between the affective cues of the two modalities to infer whether a video is ``real'' or ``fake''.  We evaluated our method on two benchmark audio-visual deepfake datasets, DFDC, and DF-TIMIT.

Our approach has some limitations. First, our approach could result in misclassifications on both the datasets, as compared to the one in the real video. Given different representations of expressing perceived emotions, our approach can also find a mismatch in the modalities of real videos, and (incorrectly) classify them as fake.  Furthermore, many of the deepfake datasets primarily contain more than one person per video. We may need to extend our approach to take into account the perceived emotional state of multiple persons in the video and come with a possible scheme for deepfake detection.

In the future, we would like to look into incorporating more modalities and even context to infer whether a video is a deepfake or not. We would also like to combine our approach with the existing ideas of detecting visual artifacts like lip-speech synchronisation, head-pose orientation, and specific artifacts in teeth, nose and eyes across frames for better performance. Additionally, we would like to approach better methods for using audio cues.  
% Since our approach is complementary to other deepfake detection methods, in the future, we would like to integrate our approach with such methods.

% \do not clearpage

%% file: sections/07-acknowlwdgements.tex
\section{Acknowledgements}
This  research  was  supported  in  part  by  ARO  Grants W911NF1910069 and W911NF1910315 and Intel.